\documentclass[letterpaper]{article}

\usepackage{natbib,alifeconf}  
\usepackage[hidelinks]{hyperref}
\usepackage{amsmath}
\usepackage{url,cleveref}
\usepackage{booktabs}
\usepackage{graphicx} 
\usepackage[shortlabels]{enumitem}
\usepackage{todonotes}

\usepackage{verbatim}

\usepackage{blindtext}
\usepackage{bbding} 
\newcommand{\cites}[1] {\citeauthor{#1}'s~(\citeyear{#1})}

\makeatletter
\renewcommand{\paragraph}{%
  \@startsection{paragraph}{4}%
  {\z@}{1.75ex \@plus 1ex \@minus .2ex}{-1em}%
  {\normalfont\normalsize\bfseries}%
}
\makeatother

\newcommand\blfootnote[1]{%
  \begingroup
  \renewcommand\thefootnote{}\footnote{#1}%
  \addtocounter{footnote}{-1}%
  \endgroup
}

\title{Visualising the Attractor Landscape of Neural Cellular Automata}

\author{
    James Stovold$^{1,}$\textsuperscript{\textdagger,*}, 
    Mia-Katrin Kvalsund$^{2,}$\textsuperscript{\textdagger}, 
    Harald Michael Ludwig$^3$, \\
    {\Large Varun Sharma$^4$ \and
    Alexander Mordvintsev$^5$ }\\
    \mbox{}\\
    $^1$University of York, UK,
    $^2$University of Oslo, Norway \\
    $^3$Machine Learning Research Unit, TU Wien, Austria,
    $^4$CeMM Vienna, Austria \\
    $^5$Paradigms of Intelligence Team, Google, Zurich \\
    \url{james.stovold@york.ac.uk}
}

\begin{document}

\maketitle

\begin{abstract}
 As Neural Cellular Automata (NCAs) are increasingly applied outside of the toy models in Artificial Life, there is a pressing need to understand how they behave and to build appropriate routes to interpret what they have learnt. By their very nature, the benefits of training NCAs are balanced with a lack of interpretability: we can engineer emergent behaviour, but have limited ability to understand what has been learnt.

 In this paper, we apply a variety of techniques to pry open the NCA black box and glean some understanding of what it has learnt to do. We apply techniques from manifold learning (principal components analysis and both dense and sparse autoencoders) along with techniques from topological data analysis (persistent homology) to capture the NCA's underlying behavioural manifold, with varying success.

 Results show that when analysis is performed at a macroscopic level (i.e.\ taking the entire NCA state as a single data point), the underlying manifold is often quite simple and can be captured and analysed quite well. When analysis is performed at a microscopic level (i.e.\ taking the state of individual cells as a single data point), the manifold is highly complex and more complicated techniques are required in order to make sense of it.
\end{abstract}


\noindent{}Data/Code available at: \url{https://github.com/jstovold/ALIFE2026}
\blfootnote{\textcopyright 2026 Stovold et al. Published under a Creative Commons Attribution 4.0 International (CC BY 4.0) license. \\ 
* Corresponding author \\ 
\textdagger{} These authors contributed equally to this work.}

\section{Introduction}

How do you know what your NCA has learnt? We typically rely on observation and argument to demonstrate that our 
NCA has learnt something. When used as a toy model of morphogenesis in artificial life, the unpredictable 
behaviour of NCAs was often considered a feature, rather than a bug. Recent advances have helped to increase the 
stability of the attractor space~\citep{stovold_identityincreasesstability}, but there remains limited 
understanding of why it might not behave as expected.

As potential application areas for learnable self-organising systems increase---for example, in computational 
biology~\citep{hartl_neuralcellularautomata}, as models of the neocortex~\citep{kvalsund_sensormovementdrives}, 
or as the basis for plastic computing platforms~\citep{barandiaran_growingreservoirsdevelopmental}---there is a 
pressing need to start building confidence that NCAs will behave the way we expect.

While the formal methods field is attempting to construct a bottom-up solution to this problem by verifying 
neural networks~\citep{attala_modellingverifyingrobotic,brucker_verifyingfeedforwardneural}, the explainable AI 
field is using a top-down approach to pick apart what a neural network has learnt (e.g.\ via sparse 
autoencoders~\citep{bricken_towardsmonosemanticitydecomposing}). For some discrete, low dimensional problems 
(e.g.\ some genetic algorithms), fitness landscapes can be visualised using local optima 
networks~\citep{ochoa_studyNKlandscapes, ochoa_localoptimanetworks} and search trajectory 
networks~\citep{ochoa_searchtrajectorynetworks_j}, but only minimal attention is paid to continuous 
landscapes~\citep{adair_localoptimanetworks}.

In this paper, we take a top-down view and analyse an NCA's behaviour to visualise its learnt attractor landscape. 
Through this approach, we can observe how the NCA state is changing as it updates, as well as how the attractor 
landscape forms at different stages of training. These visualisations are bolstered by quantitative analysis 
(using techniques from topological data analysis) and provide useful insights into the training process, helping 
to build confidence in what the NCA has learnt.


\begin{figure*}[h!]
 \centering
 \includegraphics[width=0.9\linewidth]{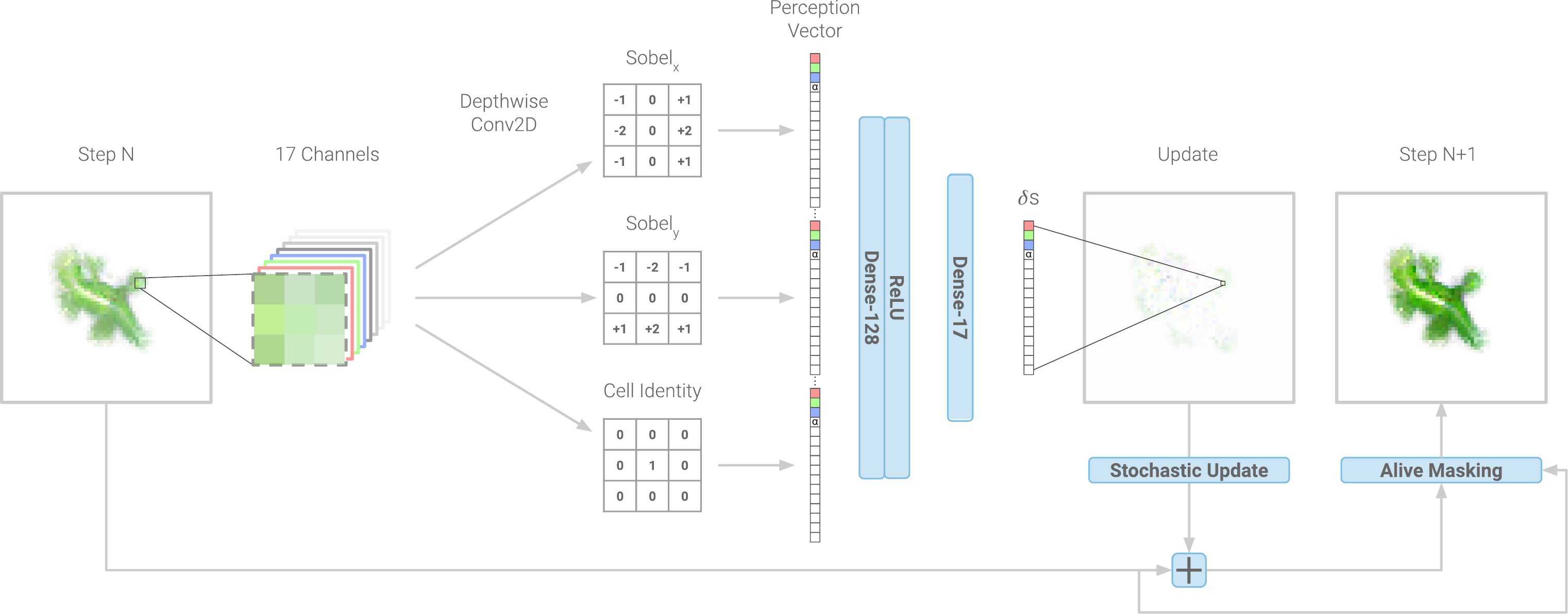}
 \caption[]{Diagram depicting one pass of the NCA update step used in this paper. The diagram also shows the structure of the neural network. Image adapted from \citep{mordvintsev_growingneuralcellular}, licenced under CC BY 4.0.}
 \label{fig:nca_update}
\end{figure*}

\section{Neural Cellular Automata}

The Neural Cellular Automata (NCA) is an extension to the classic cellular automata 
model~\citep{vonneumann_theoryselfreproducing}, which consists of a grid of cells (typically one- or 
two-dimensional) where each cell can be in one of a finite number of states (originally just 0 or 1). An 
automaton is defined that, at each timestep, updates the state of each cell based on the state of its 
neighbouring cells~\citep{izhikevich_gameoflife}.

The NCA~\citep{mordvintsev_growingneuralcellular} replaces the discrete cell state with a vector of reals---usually 16 or 17 `channels', where the first 4 are used to visualise the state by treating them as RGBA values. 
The automaton is replaced by a small neural network~\citep{li_calibrationcellularautomata} and convolutions are 
applied for efficient neighbourhood calculations~\citep{gilpin_cellularautomataconvolutional}. 
Fig.~\ref{fig:nca_update} shows one update step of the NCA.

\begin{figure}[h!]
 \includegraphics[width=0.3\linewidth]{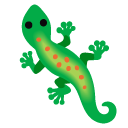}
 \includegraphics[width=0.3\linewidth]{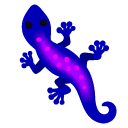} \quad 
 \includegraphics[width=0.3\linewidth]{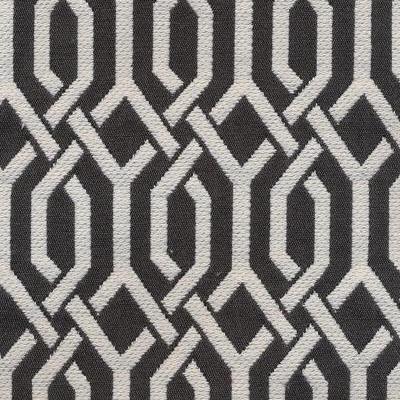}

 \caption[]{Example classic target images used for training NCA models. Left, a green gecko used 
 in~\citep{mordvintsev_growingneuralcellular}; middle, a blue gecko used in~\citep{stovold_ncascanrespond}; 
 right, `interlaced 0172' texture from~\citep{cimpoi_describingtextureswild} used 
 in~\citep{niklasson_selforganisingtextures}.}

 \label{fig:example}
\end{figure}

NCAs are generally employed in one of two ways: those trained to produce a single pattern (such as the classic 
gecko shape, shown in fig.~\ref{fig:example} (left)), and those trained on textures (such as the image shown in 
fig.~\ref{fig:example} (right)), which generally display more dynamic behaviour but have regular structure.

Just as ordinary differential equations (ODEs) can be parameterised with neural networks, and make use of the 
backpropagation algorithm to train them (in the form of `neural ODEs'~\citep{chen_neuralordinarydifferential}), 
the NCA can be thought of as the equivalent for some forms of partial differential equations~\citep{mordvintsev_differentiableprogrammingreaction}. The NCA is learning a parameterised variant of 
the general reaction--diffusion equations:

\begin{equation}
  \frac{\partial{\mathbf{q}}}{\partial{t}} = R_\theta(\mathbf{q}, D_\phi(\mathbf{q}))
\end{equation}

\noindent{}where $\mathbf{q}$ is the state of the system, $R_\theta$ is the parameterised reaction/update 
function that depends on the current state $\mathbf{q}$, and $D_\phi(\mathbf{q})$ is the diffusion/influence 
from neighbouring cells (the `perception vector'). Training the system consists of learning $\theta$ using 
backpropagation-through-time until the system meets some target behaviour (e.g.\ grows into a 
shape within a set number of timesteps). Depending on the type of NCA, the diffusion term can be learnt as 
well (parameterised by $\phi$) which relates to letting the system learn the convolutional kernels, as 
in~\citep{randazzo_selfclassifyingmnist}.

For the classic gecko growth example, $\mathbf{q}$ starts as a single seed cell in an empty environment, $\phi$ 
is fixed (the convolutional filters are not trained), and $\theta$ is trained. The loss is calculated by 
comparing the state $\mathbf{q}$ after 64--96 timesteps to the RGB image of the green gecko in 
fig.~\ref{fig:example} (left), using root mean squared error (RMSE). 


Parameterising the system in this manner gives us one clear advantage: with the backpropagation 
algorithm~\citep{rumelhart_learningrepresentationsbackpropagating}, we gain the ability to engineer the system's 
behaviour in a top-down manner, something notoriously difficult for self-organising 
systems~\citep{stepney_engineeringemergence}. In doing so, however, we lose the interpretability of the 
model---instead of having an equation or formula, we have a black box neural network.

If we wanted to ask what a particular NCA has learnt (or, similarly, whether our NCA has learnt what we think it 
has learnt), we have two approaches: we can either (a) try to extract the equations 
$\tfrac{\partial{\mathbf{q}}}{\partial{t}}$ etc.\ from $\theta$ and check analytically that they correspond to 
what we think it has learnt (highly non-trivial); or (b) we can observe the behaviour of the system and attempt 
to construct some underlying behavioural manifold that allows us to see the attractors that have been learnt and 
their relative placement in some latent space.

In this paper, we opt for the second approach. We apply a variety of techniques to visualise and analyse the 
attractor space learnt by the NCA. Most of these techniques have been drawn from neuroscience, 
single-cell biology, or from the explainable AI field. We show that for some cases, the attractor space of an 
NCA can be visualised and analysed. We show both successful and unsuccessful examples to help guide those who deploy these techniques.

\section{Methods}

\subsection{Datasets}The datasets used in this paper are extracted from two NCA models: the blue/green gecko model (where the gecko 
changes colour when it receives a signal from the environment~\citep{stovold_ncascanrespond}), and the interlaced texture 
model. We also briefly test our approaches using data from \cites{mordvintsev_emergentmultiscalestructures} Haeckel NCA. We extract behavioural data from these models by running the NCA from an initial seed state and recording 
the state of the NCA at each timestep. For the blue/green gecko, we ran the model for 20,000 timesteps, and 
provided the signal to change colour every $150$ timesteps; the signal was not provided to exactly the same 
place each time, there was some jitter around a central point. The environment was $60\times 60$ with $17$ 
channels. For the texture data, we extracted $250$ timesteps of $192\times 192$ with $12$ channels.

\subsection{Techniques Applied}

\paragraph{Manifold Learning} The manifold hypothesis~\citep{fefferman_testingmanifoldhypothesis} posits that 
complex systems whose states exist in a high-dimensional space are likely to occupy a lower-dimensional subspace 
rather than exploring the entire volume of high-dimensional space. This subspace may have a coordinate system 
that is more informative than the original coordinates of the high-dimensional dataset. The geometric and 
topological properties of this subspace may possess relevant information about the properties of the 
system-level behaviour.


Manifold learning~\citep{lee_nonlineardimensionalityreduction} is the process of applying dimensionality 
reduction techniques to extract this low-dimensional manifold directly from the high-dimensional data, with the 
aim of representing the manifold's properties as accurately as possible for human interpretation.

\paragraph{Autoencoders}

One such dimensionality reduction technique is an autoencoder. The 
autoencoder~\citep{rumelhart_learningrepresentationsbackpropagating} is a neural network trained to reconstruct 
its input through a bottleneck. The network consists of an encoder, which maps the input to a lower-dimensional 
latent representation, and a decoder, which attempts to reconstruct the original input from this compressed 
code. By minimising reconstruction error, the autoencoder is forced to learn a compact representation that 
captures the most salient structure in the data.

When the encoder and decoder are both single linear layers and the loss is mean squared error, the autoencoder 
recovers a subspace equivalent to that found by Principal Component Analysis (PCA)~\citep{baldi1989neural}. In 
this sense, PCA can be understood as a special case of the autoencoder framework. By introducing nonlinear 
activation functions and deeper architectures, autoencoders can capture structure that PCA cannot, for example, 
curved or folded manifolds that a linear projection would collapse~\citep{hinton2006reducing}.

For our purposes, in our macroscopic analysis we use a convolutional autoencoder to project the $60\times 60 \times 17$ 
NCA state into a 2-dimensional latent space suitable for visualisation. Each point in the resulting latent space 
corresponds to a full NCA grid state at a particular timestep. We colour these points according to 
properties of the original state (for example, the average pixel colour) to track the system's trajectory 
through its behavioural space. For our microscopic analysis we use a simpler non-convolutional autoencoder to 
project the 17-dimensional cell states into a 2-dimensional latent space, and colour the data point with the RGB 
value of the original data point.

The sparse autoencoder (SAE) differs from a standard autoencoder in two important 
respects: (a) rather than compressing the input into a lower-dimensional bottleneck, the SAE projects into a 
higher-dimensional latent space~\citep{olshausen1997sparse}; (b) a sparsity penalty on the latent 
activations encourages the network to use only a small number of these features for any given input. Where a 
dense autoencoder learns a distributed, entangled code useful for visualisation, an SAE tends to produce more 
interpretable features with each latent dimension more likely to correspond to a distinct, identifiable property 
of the input. This aligns with recent work in mechanistic interpretability, where SAEs have been used to 
decompose the internal representations of large language models into monosemantic 
features~\citep{bricken_towardsmonosemanticitydecomposing}. In our setting, SAEs offer a route toward 
understanding what aspects of the NCA state each latent dimension is encoding, rather than simply providing a 
projection for visualisation.


\paragraph{Persistent Homology}

Dimensionality reduction tools are helpful when trying to visualise manifolds from higher dimensions in two or 
three dimensions. When seeking to apply these tools, it can be difficult to know whether the real manifold is 
captured. PCA, for example, will not fare well with a folded manifold, as it will compress the manifold 
together along the dimensions of linear projection. It can therefore be useful to have an approximate ground truth 
when using these techniques, to verify or understand the projections.

\begin{figure*}
    \centering
    \includegraphics[width=\linewidth]{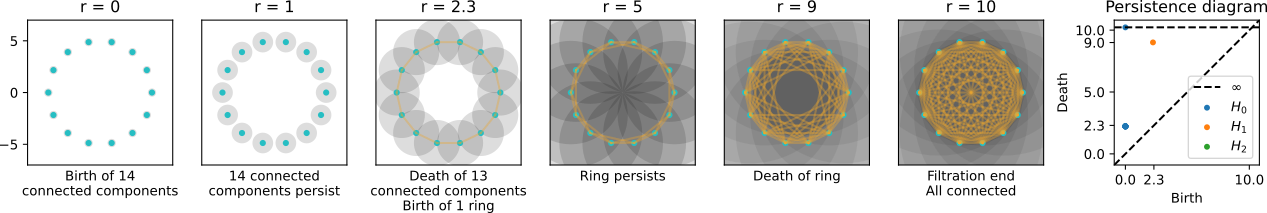}
    \caption{The creation of the persistence diagram (right) consist of growing circles around your points with 
    increasing radius $r$, and connecting components if the growing circles overlap with a point. Here, the toy 
    data is in blue, growing cirles are in shaded grey, and edges between connected components are in orange.}
    \label{fig:PH_explanation}
    \vspace{-0.5em}
\end{figure*}

Persistent Homology (PH) seeks to provide such an approximate ground truth by finding topologically invariant features in the 
raw high-dimensional data~\citep{zomorodian2004computing, carlsson2009topology}. This analysis is robust to both small perturbations and continuous transformations 
such as folding and stretching in higher dimensions~\citep{cohen2005stability, carlsson2009topology}. We provide a brief introduction to the method here, but the reader is guided to \citep{kemme2025persistent} for further details.

The topological invariants of interest are primarily:
\begin{itemize}[noitemsep]
    \item $H_0$: Connected components, isolated objects
    \item $H_1$: Rings / loops / cycles
    \item $H_2$: Voids
\end{itemize}

These features are found by iteratively constructing simplicial complexes with an increasing Euclidean distance 
$r$ (see fig.\ \ref{fig:PH_explanation} for an example).  For a complex created at radius $r$, a set amount of 
topological features are detected. These features appear at a given $r$ as components become connected, and 
disappear at a higher $r$ as features are merged or holes are filled. This continuous filtration is handled by 
the \texttt{ripser} library~\citep{ctralie2018ripser}, and gives us a persistence diagram. Significant features 
can be seen as any feature that persists for many radii. These can be seen in the diagram as any feature far off 
the diagonal.

A high-dimensional manifold can be recognised with PH analysis by its Betti number\footnote{commonly denoted with 
$\beta_i$, we simply refer to the homology groups $H_i$ here to simplify the terminology}~\citep{carlsson2009topology}. For example, if 
counting one connected component and one void, these Betti numbers correspond to a sphere. If seeing two 
connected components and two voids, we might equally infer we are seeing two independent spheres. See table 
\ref{tab:betti} for example Betti numbers.

\begin{table}
    \centering
    \begin{tabular}{l|c|c|c}
        & $H_0$ & $H_1$ & $H_2$ \\
        \hline
        Ring & 1 & 1 & 0 \\
        Two disjoint rings & 2 & 2 & 0 \\
        Sphere & 1 & 0 & 1 \\
        Torus & 1 & 2 & 1 \\
    \end{tabular}
    \caption{Betti numbers for common topologies. In general, $H_0$ tells us how many distinct components there 
    are in the data, $H_1$ tells us how many rings exist, and $H_2$ tells us how many voids there are. Adapted from \citep{kemme2025persistent}.}
    \label{tab:betti}
    \vspace{-1em}
\end{table}

Though robust, PH does have its limitations. Primarily, topological objects can be quite reductive, as an object 
such as a cup and a donut are seen as topologically equivalent~\citep{kemme2025persistent}. While higher-dimensional features are possible to compute, computing features $H_3$ and onwards in practice are quite time- and memory-intensive. 
One of the main assumptions for dimensionality reduction, however, is that most manifolds in 
high-dimensional space will be meaningfully low-dimensional. If no $H_1$ or $H_2$ groups are found with PH, a 
higher dimensional feature could be present but not detected. In practice, a more likely explanation is that the manifold simply has no holes, or that the sampling is poor or noisy.

\begin{figure}[h!]
    \centering
    \includegraphics[width=0.9\linewidth]{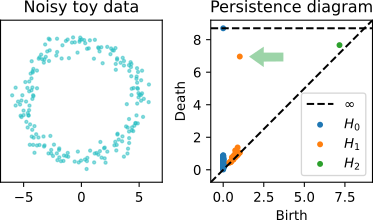}
    \caption{Persistence diagram for noisy data. The green arrow points to a strongly persistent ring. The rest 
    of the recorded components, like the $H_2$ group that briefly appears at $r = 7.1$, can be considered noise 
    because their persistence is short-lived.}
    \label{fig:PH_explanation_noisy}
    \vspace{-0.5em}
\end{figure}

Two specific problems encountered when dealing wih NCAs are that of noise and subsampling. First, as can be seen 
in fig.\ \ref{fig:PH_explanation_noisy}, real data will yield less clean PH diagrams. Second, NCA substrates 
tend to be quite large, and run for many timesteps, generating a lot of datapoints. To use PH libraries, 
the data will need to be subsampled. If the coverage is poor, the underlying manifold may not be accurately 
captured; and loops and voids that are not sufficiently persistent might be detected and appear as noise. 
Therefore, as we will see in the Results section, some NCA data pose problems for PH analysis.

\section{Results}

In this section we present results from analysing NCAs at two levels: microscopic and macroscopic. The 
microscopic analysis considers the NCA from the perspective of the individual cell/pixel, where each cell is 
acting based on its neighbourhood and will find the most stable state it can given its context. The macroscopic 
analysis considers the NCA from the perspective that the grid of cells is a single data point, and is more 
aligned to how NCAs are trained.

\subsection{Macroscopic Analysis}

Consider the gecko NCA trained by \citet{stovold_ncascanrespond} to change colour from green to blue when it 
receives a signal. \citet{stovold_ncascanrespond} claims that the external signal pushes the system between the 
same two attractors (rather than pushing the system along a finite series of attractors). This was demonstrated 
by repeatedly providing the signal many more times (i.e.\ $10\times$) than the NCA was exposed to during training. In 
this section, we use this model to demonstrate how the aforementioned techniques can be applied to visualise the 
attractor landscape, not only during operation but also at different stages of the training process.

\begin{figure*}[h!]
 \centering
 \includegraphics[width=0.7\linewidth]{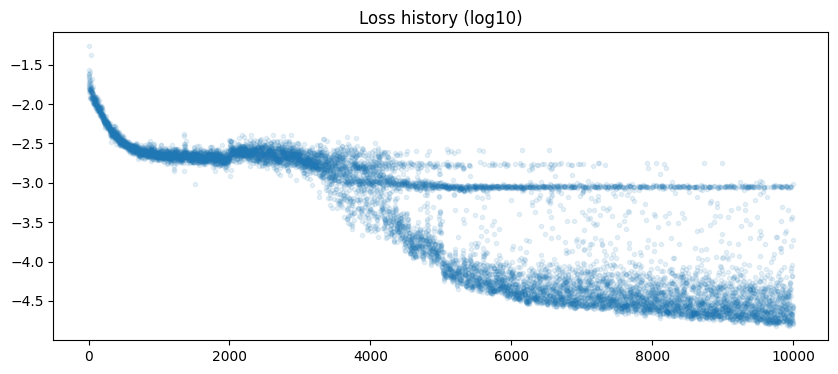} \\
 \includegraphics[width=0.9\linewidth]{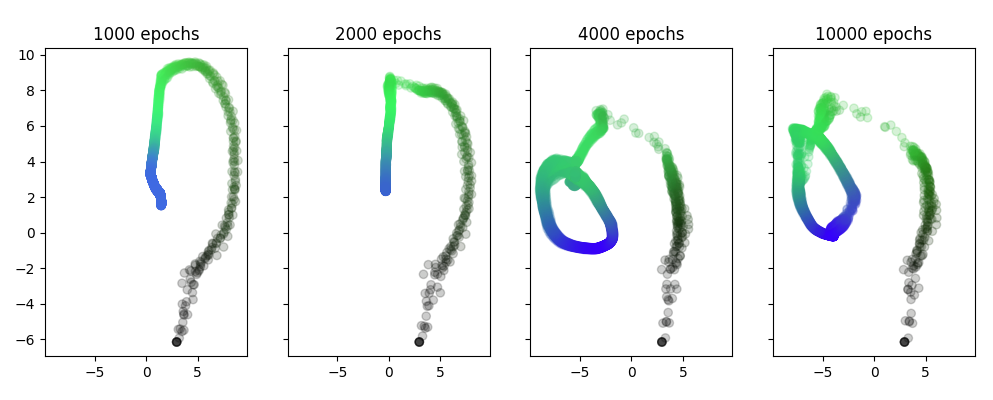} \\
 \includegraphics[width=0.9\linewidth]{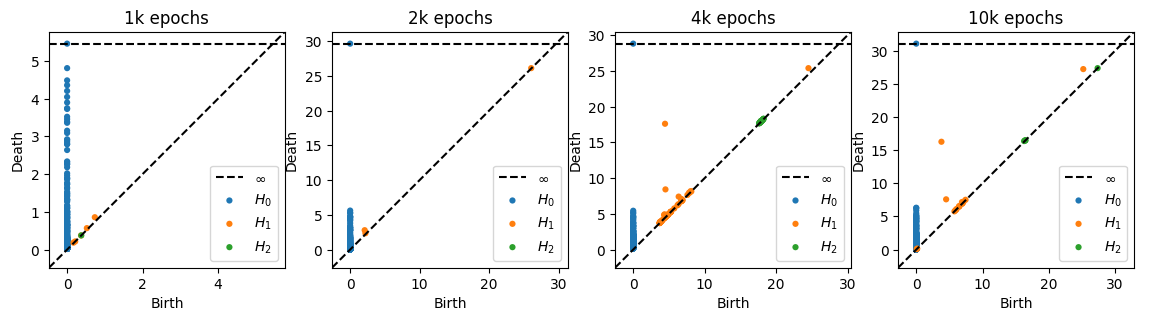} 

 \caption[]{ The blue/green gecko model from \citep{stovold_ncascanrespond} being trained. (Top) loss as it 
 trains, showing transition at around 3500--4000 epochs; (middle) latent space of the autoencoder model after 
 training for the 1000, 2000, 4000, and 10,000 epochs; (bottom) persistent homology of the underlying manifold at 
 the corresponding training points, confirming a cycle forms ahead of 4000 epochs.}

 \label{fig:4models}
 \vspace{-1em}
\end{figure*}

Fig.\ \ref{fig:4models} shows how the autoencoder and persistent homology techniques can be applied to better 
understand what the NCA model has learnt. The top image shows a typical loss log, with a shift in behaviour at 
around 4000 epochs; from this we can deduce that it learnt to do something particularly important at around this 
time. The middle image shows the latent space of the autoencoder, after having trained on many examples of the 
NCA's behaviour at the specified training epochs; each point in the image is coloured according to the average 
colour of all pixels in the NCA at that point.

The images on the left of the middle row (1000 and 2000 epochs) show that the system has learnt to grow from a 
single seed (which is at the bottom of each sub-image) into a green gecko (top), and then changes colour to a 
blue gecko (middle left). After the phase transition at around 4000 epochs, we can see something qualitatively 
different in the latent space: in the images on the right of the middle row (4000 and 10,000 epochs) there is now 
a cycle between the green and blue gecko. This suggests that, during the shift in behaviour observed around 4000 
epochs, the NCA has learnt to change its colour back to a green gecko.

Whether this is the {\em same} green gecko is not guaranteed at this stage. For example, if the autoencoder 
happened to have compressed the states into the two-dimensional representation in such a way that the green 
states were all stacked on top of each other (which would require another dimension for us to see) then we would 
not have a cycle here, but instead just the top-down view of a helix.

To confirm that this really is a cycle, we calculated the persistent homology of the dataset. The bottom image 
in fig.~\ref{fig:4models} shows the output of this. For 1000 and 2000 epochs, there are no large cycles in the 
manifold, but for 4000 and 10,000, a single large cycle is present and persists. This confirms that the rhetoric 
of \citet{stovold_ncascanrespond} about the system returning to the {\em same} gecko is correct.

\paragraph{Robustness to perturbation}

\begin{figure}[h!]
 \includegraphics[width=\linewidth]{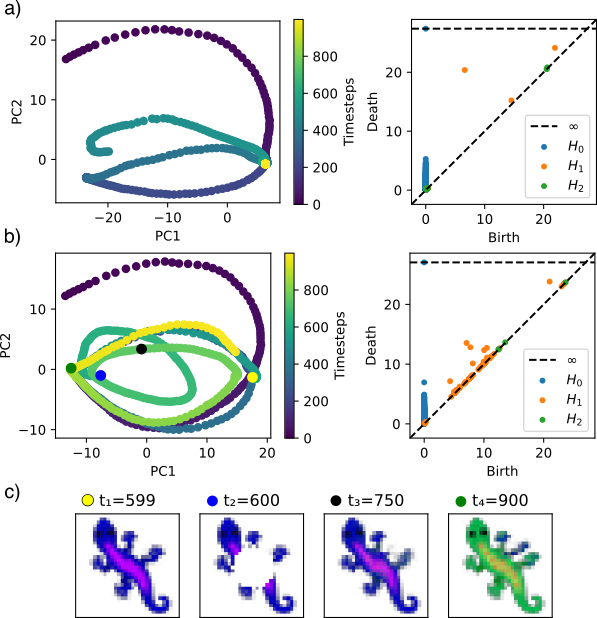}

 \caption[]{Perturbation on the blue/green gecko model. The left graphs show the first two principal components 
 of the dataset, the right graphs show the persistent homology analysis. (a) A perturbation was applied on the 600th timestep, as the system was in the blue attractor. It returns to the green attractor and stays there until the episode ends. (b) The same was done, but the signal to change colour was still sent every 150 timestep. (c) State of the NCA at different timesteps.}

 \label{fig:perturbation}
 \vspace{-1em}
\end{figure}

Many NCAs are trained to be robust to perturbation, able to `heal' themselves after exposure to damage. In this 
section, we visualise this healing process in the attractor landscape. Using a similar approach to above, we can 
ask whether the system returns to the same attractor, and whether the system really has stabilised. Fig.\ 
\ref{fig:perturbation}(c) shows the dataset we are working with: we take the blue/green gecko model and perturb 
the system after 600 timesteps (by setting the pixels in the middle of the image to 0). After this 
perturbation the system returns to normal behaviour and still changes colour when signalled.

When a system is robust, a perturbation will return the state of the system back to the attractor and the PH 
will show the same as before: one loop. Fig.\ \ref{fig:perturbation}(a) shows that this is the case, the system 
returns to an attractor and the PH analysis shows us that the ring persists.\footnote{Due to topological 
reduction, a ring with a path sticking out is equivalent to a ring}

\begin{figure}[h!]
 \centering
 \includegraphics[width=0.9\linewidth]{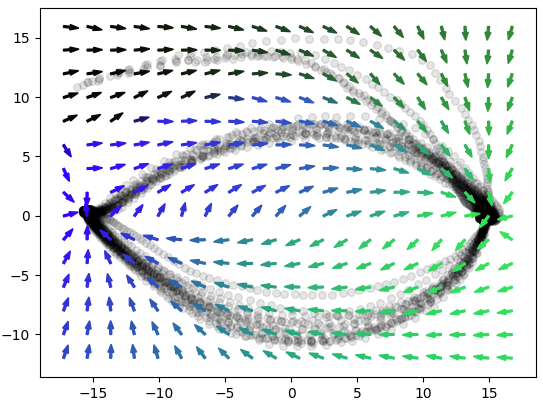}

 \caption[]{Field lines superposed on the underlying latent space (captured using PCA). To visualise the vector 
 field, we sampled this latent space, reconstructed the high-dimensional representation, ran it through the NCA 
 for a small number of steps and then reduced it back down to the latent space. This allows us to capture which 
 direction the NCA model would move in the latent space, and hence to visualise the behaviour of the system 
 across the latent space.}

 \label{fig:field_lines}
\end{figure}

What is noteworthy, however, is that it has returned to the wrong attractor. While the perturbation pushed the 
system away from the blue attractor, it returned to the green. By mapping the field lines (fig.\ 
\ref{fig:field_lines}), we can reason about why this has happened and about its ability to return to the correct 
attractor: observe that a perturbation towards the top of the latent space will always go towards the green 
attractor and perturbations towards the bottom of the latent space will always go towards blue.

If we continue to apply the external signal (to change colour) after a perturbation, we can follow what happens 
to the system during its return to an attractor. In this scenario, the system doesn't make it fully back to the 
attractor before the signal is provided, meaning that it spirals between the two attractors (fig.\ 
\ref{fig:perturbation}(b)), and the PH analysis shows that there are now several rings present.


\subsection{Microscopic Analysis}

\begin{figure}[h!]
 \centering
 \includegraphics[width=0.9\linewidth, trim={0 0 15mm 0}, clip]{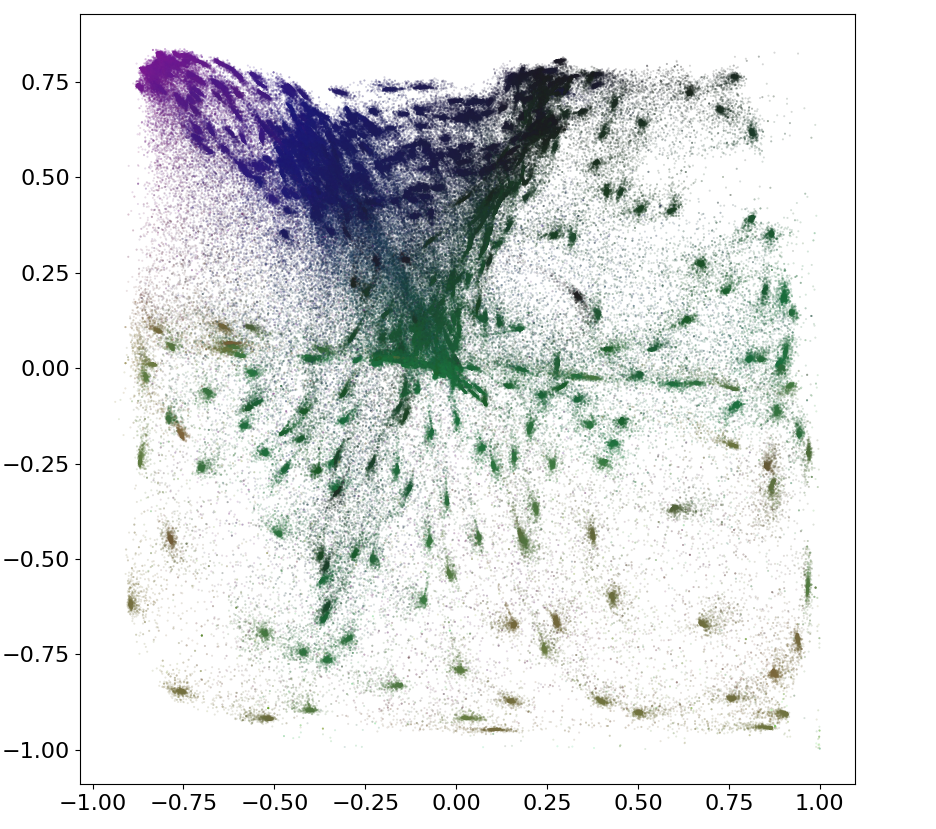}

 \caption[]{A sample of 1 million data points (excluding dead cells) encoded using an autoencoder to visualise 
 the underlying latent space. The clumps of cells indicate local attractors for individual cells. The NCA model 
 used was the 10,000 epoch blue/green gecko model. } 

 \label{fig:bluegreen_pixel}
 \vspace{-1em}
\end{figure}

In contrast to the clean, easily-interpretable attractor landscapes we see at the macroscopic level, at the 
microscopic level we get a much messier picture. Fig.\ \ref{fig:bluegreen_pixel} shows the latent space of the 
same blue/green gecko dataset when interpreted at an individual pixel level. There is clearly structure visible 
here, but the number of apparent attractors is much larger; each clump of pixels in this space appears to 
indicate an attractor, but which particular attractor the pixel lands in depends on the neighbourhood around it.


The reason the space grows more complex compared to the relatively simple attractor landscape at macroscopic 
level is due to the NCA being trained to produce simple dynamics at a macroscopic level (i.e.\ the loss is 
calculated at that level). We also have much more data to work with: for $T$ timesteps, at macroscopic level we had $T$ images to analyse; at microscopic level we have $T\times N \times M$ where $N\times M$ is the size of the NCA, which can make it difficult to compute and ascertain the persistent homology.

While the microscopic attractor landscape for the gecko is highly complex, a regular pattern should be 
comparatively simple. if we have some regularity at the macroscopic level, microscopic analysis can still be 
useful.

Consider the texture NCA seen in fig.\ \ref{fig:example} (right). We collected $250$ timesteps of a 
$192\times192$ NCA, giving us a total of ca.\ 9 million datapoints. It is reasonable to assume that if the 9 
million pixels sit on any manifold, the regularity of the image would necessitate that any sufficiently-large 
subsample of the image would show the same manifold. 

We subsample the dataset, preserving the pixel locality (i.e.\ we take a window of size $20\times20$ of the wider NCA and analyse 
that window across all timesteps). This should let us get a more complete sample of the manifold with less 
noise.



\begin{figure}[h!]
 \centering
 \includegraphics[width=\linewidth]{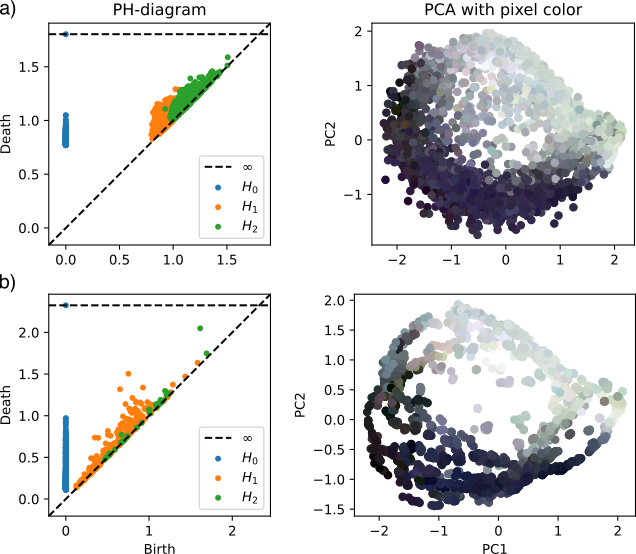}

 \caption[]{The PH diagrams (left) and the PCA projection of a) the full substrate and b) the subsampled 
 substrate, based on 20 timesteps.}
 
 \label{fig:texture_ph}
 \vspace{-1em}
\end{figure}

In fig.\ \ref{fig:texture_ph}, we see a comparison between analysing the full image and the subsample for 
timesteps 100 to 120 (after the transients have subsided). Although the PCA looks fairly similar, the PH diagram 
is very noisy for the full image and shows no significant objects. The subsample, meanwhile, reveals at least 
two persistent loops and one persistent void, indicating that we are likely dealing with a torus. This is only 
revealed when the PH analysis gets decent coverage of the data (12.5\% for the subsample compared to 0.136\% of the full dataset).


\paragraph{Sparse feature decomposition}

The dense autoencoder projection in fig.~\ref{fig:bluegreen_pixel} shows that individual cells cluster into 
local attractors, but the resulting latent space is difficult to interpret: we see structure without knowing 
what distinguishes the clumps. To probe this further, we trained a sparse autoencoder on the 17-channel cell 
states of the blue/green gecko model, using a dictionary of $1088$ features (a $64\times$ expansion of the input 
dimension) with an $L_1$ sparsity penalty. The trained SAE reaches a reconstruction error below $10^{-6}$ while 
using only 5--8 active features per cell, with roughly $90\%$ of dictionary features dead.

\begin{figure}[h!]
 \centering
 \includegraphics[width=0.8\linewidth]{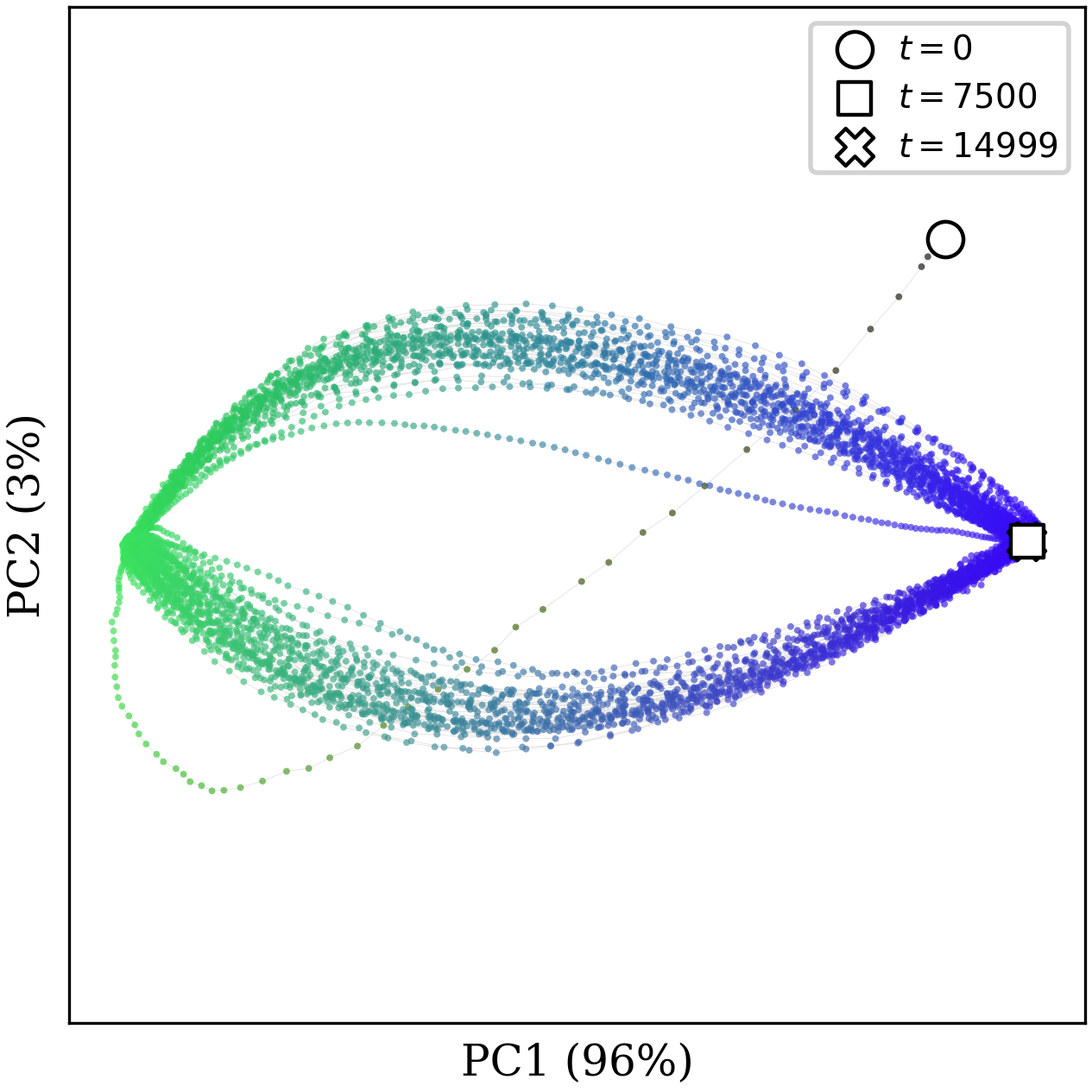}
 \caption[]{Per-frame manifold of the blue/green gecko NCA in SAE feature space. Each point is the mean SAE activation vector across all living cells in one frame, projected using PCA and coloured by the mean RGB value of the frame. The two attractors sit at opposite ends of the first principal plane and the trajectory traces a clean cycle between them, recovering the macroscopic cycle of fig.~\ref{fig:4models} from purely microscopic per-cell data.}
 \label{fig:sae_manifold}
 \vspace{-1em}
\end{figure}

For each frame we compute a single feature vector by averaging the SAE encodings across all living cells in that 
frame, and then project the resulting per-frame vectors using PCA. Fig.~\ref{fig:sae_manifold} shows that this 
per-frame SAE embedding recovers the same blue/green cycle that the macroscopic autoencoder found in 
fig.~\ref{fig:4models}, with the two attractors sitting at opposite ends of the first principal plane. This is a 
non-trivial result: the SAE is trained only on individual cell states, with no access to whole-grid information, 
yet averaging its sparse features per frame recovers the global attractor structure. The microscopic and 
macroscopic views, which looked incompatible under the dense autoencoder analysis, are reconciled once the cell 
states are decomposed into sparse features.

Applying the same pipeline to NCAs trained on the interlaced textures ($12$ channels) or Haeckel dataset ($24$ 
channels) yields qualitatively different outcomes. The texture NCA admits no compression: the SAE recovers 
exactly $12$ active features, essentially rediscovering the channel basis. The Haeckel NCA collapses to two 
colour-aligned features. We speculate that this is a channel-budget effect: the blue/green gecko at $17$ 
channels sits in a regime with enough representational slack for the SAE to extract compositional spatial 
features, while the texture NCA is already at its minimum description length and the Haeckel NCA's surplus 
channels have been absorbed into a redundant colour encoding.

\section{Discussion}

We have two ways of looking at an NCA and its behaviour. The NCA is trained to produce patterns at a global/macroscopic level, but the changes all take place at a local/microscopic level. This is why the global attractor space is much simpler and easier for us to comprehend, whereas the local attractor space is more complicated.

The analysis of the blue/green gecko model reflects this pattern, where the macroscopic analysis shows a simple, 
clear attractor landscape, with two attractors linked in a cycle arrangement (confirmed by the persistent 
homology analysis showing a single cycle). The microscopic analysis of this model, however, is less clear cut. 
There are regional patterns, with some commonality among them, but it is substantially harder to interpret.

Introducing multiple geckos in the same environment repeats this pattern. The techniques we have applied start to struggle as the number of possible arrangements at the global level increases. While the behaviour of each individual gecko is the same, globally we see a more complicated attractor space due to the number of different permutations. This is not dissimilar to the difference between the local, microscopic view and the macroscopic, single gecko view discussed above.


It is possible that the techniques tend to struggle with the NCA textures for the same reason. While the interlaced texture example analysed here has sufficient regularity that we are able to extract some underlying manifold, it is very noisy and requires subsampling the dataset to make it viable. We also attempted to analyse the more complex Haeckel dataset, but the problems encountered with the texture NCA were compounded by the reduced regularity.

These problems highlight the challenge of interpreting black box models, and we are looking whether clustering algorithms can capture the attractor space. For example, a Gaussian Mixture Model works reasonably well, but in doing so we lose some of the clarity we get from the aforementioned dimensionality reduction techniques, and topological data analysis is less applicable.

\section{Acknowledgements}

The authors would like to thank Kosio Beshkov for his assistance and advice on interpreting the persistent 
homology output. The authors would also like to thank the organisers of the ALICE 2026 workshop, which brought the 
team together and at which the project was initiated.

\footnotesize
\bibliographystyle{apalike}
\bibliography{paper}

\end{document}